\documentclass[10pt,twocolumn,letterpaper]{article}

\usepackage{cvpr}
\usepackage{times}
\usepackage{graphicx}
\usepackage{amsmath}
\usepackage{amssymb}

\usepackage[numbers,sort,compress]{natbib}
\usepackage{makecell}
\usepackage[dvipsnames]{xcolor}
\usepackage{booktabs}
\usepackage{authblk}

\usepackage[bookmarks=false, colorlinks=true]{hyperref}

\cvprfinalcopy 

\def\cvprPaperID{15} 

\ifcvprfinal\fi
\begin{document}

\title{Effective Data Fusion with Generalized Vegetation Index:\\ Evidence from Land Cover Segmentation in Agriculture}

\author[1]{Hao Sheng}
\author[1]{Xiao Chen}
\author[2]{Jingyi Su}
\author[1]{Ram Rajagopal}
\author[1]{Andrew Ng}

\affil[1]{Stanford University}
\affil[2]{Chegg, Inc}
{
\makeatletter
\renewcommand\AB@affilsepx{\quad \protect\Affilfont}
\makeatother

\affil[1]{\normalsize\texttt {\{haosheng,markcx,ramr\}@stanford.edu$\quad$ang@cs.stanford.edu }}
\affil[2]{\normalsize\texttt{jingyi.su.js@gmail.com}}
}



\maketitle

\begin{abstract}
How can we effectively leverage the domain knowledge from remote sensing to better segment agriculture land cover from satellite images?
In this paper, we propose a novel, model-agnostic, data-fusion approach for vegetation-related computer vision tasks.
Motivated by the various Vegetation Indices (VIs), which are introduced by domain experts, we systematically reviewed the VIs that are widely used in remote sensing and their feasibility to be incorporated in deep neural networks. To fully leverage the Near-Infrared channel, the traditional Red-Green-Blue channels, and Vegetation Index or its variants, we propose a Generalized Vegetation Index (GVI), a lightweight module that can be easily plugged into many neural network architectures to serve as an additional information input. To smoothly train models with our GVI, we developed an Additive Group Normalization (AGN) module that does not require extra parameters of the prescribed neural networks. Our approach has improved the IoUs of vegetation-related classes by $0.9-1.3$ percent and consistently improves the overall mIoU by $2$ percent on our baseline.
\end{abstract}

\section{Introduction}


Deep learning has been widely adopted in computer vision across various applications such as diagnosing medical images\cite{irvin2019chexpert}, classifying objects in photos\cite{oshri2018infrastructure}, annotating video frames\cite{karpathy2014large}, etc. However, recognizing the visual patterns in the context of agriculture, especially segmenting the multi-labeled masks, has not been explored extensively in detail. One primary reason that hinders the progress is the difficulty of handling complex multi-modal information inside the images\cite{chiu2020agriculture} because the sensing imagery in agriculture contains Near Infrared band and other thermal bands that are distinguished from traditional images spanning over red, green, and blue (RGB) visual bands. Such multi-band information is crucial for understanding the land cover context and field conditions, e.g., the vegetation of the land.

\begin{figure}[!hbpt]
\centering
\includegraphics[width=0.4\linewidth]{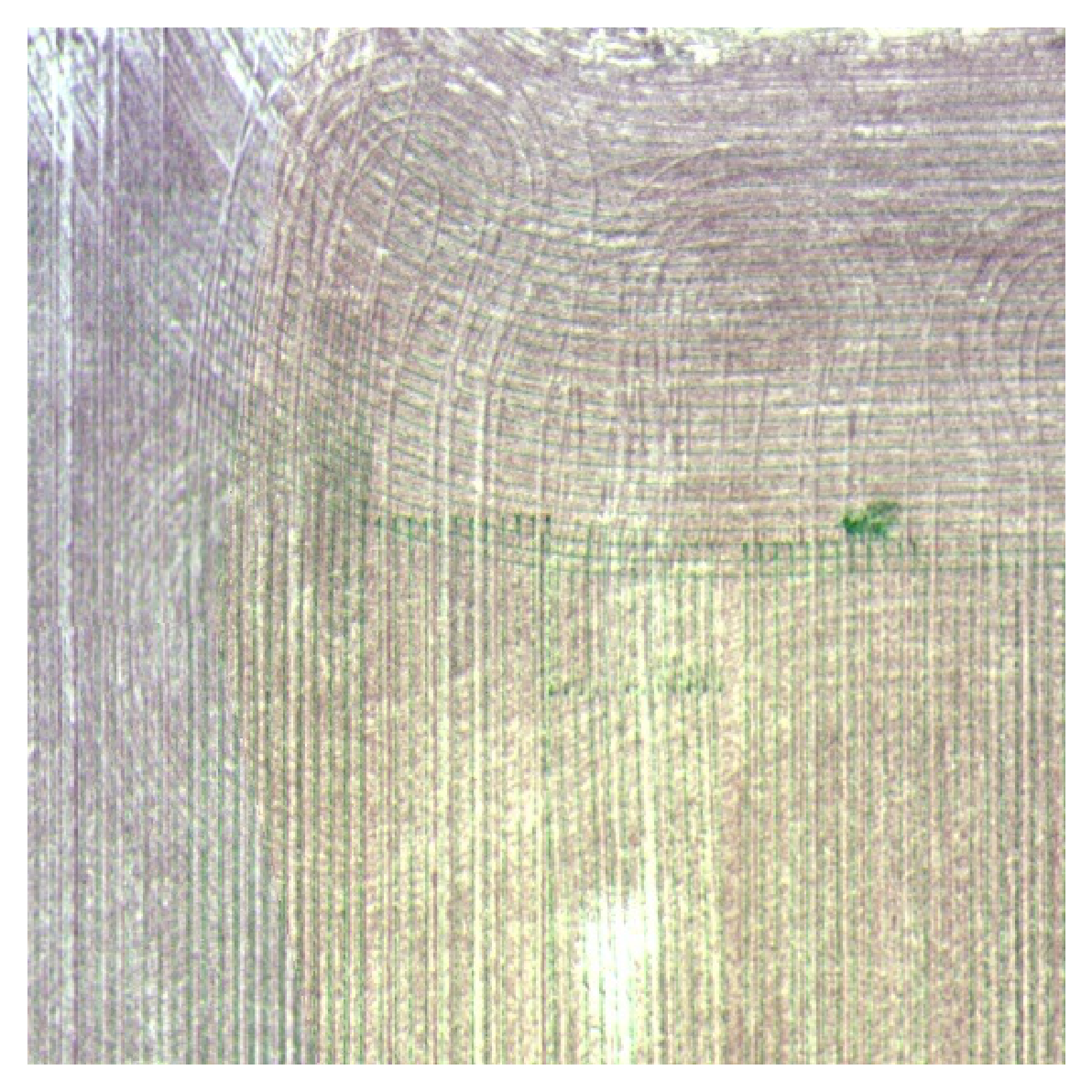}
\includegraphics[width=0.4\linewidth]{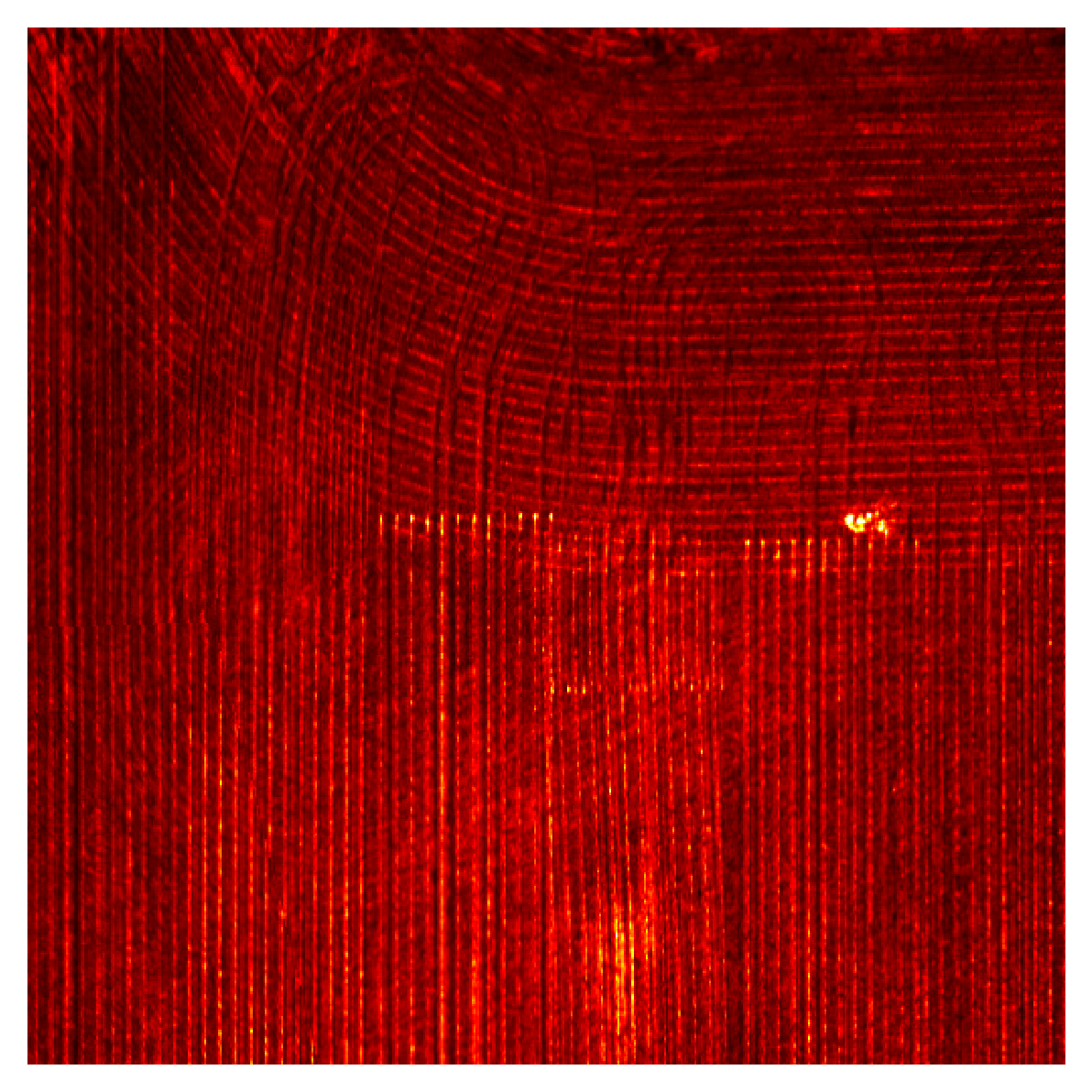}
\includegraphics[width=0.4\linewidth]{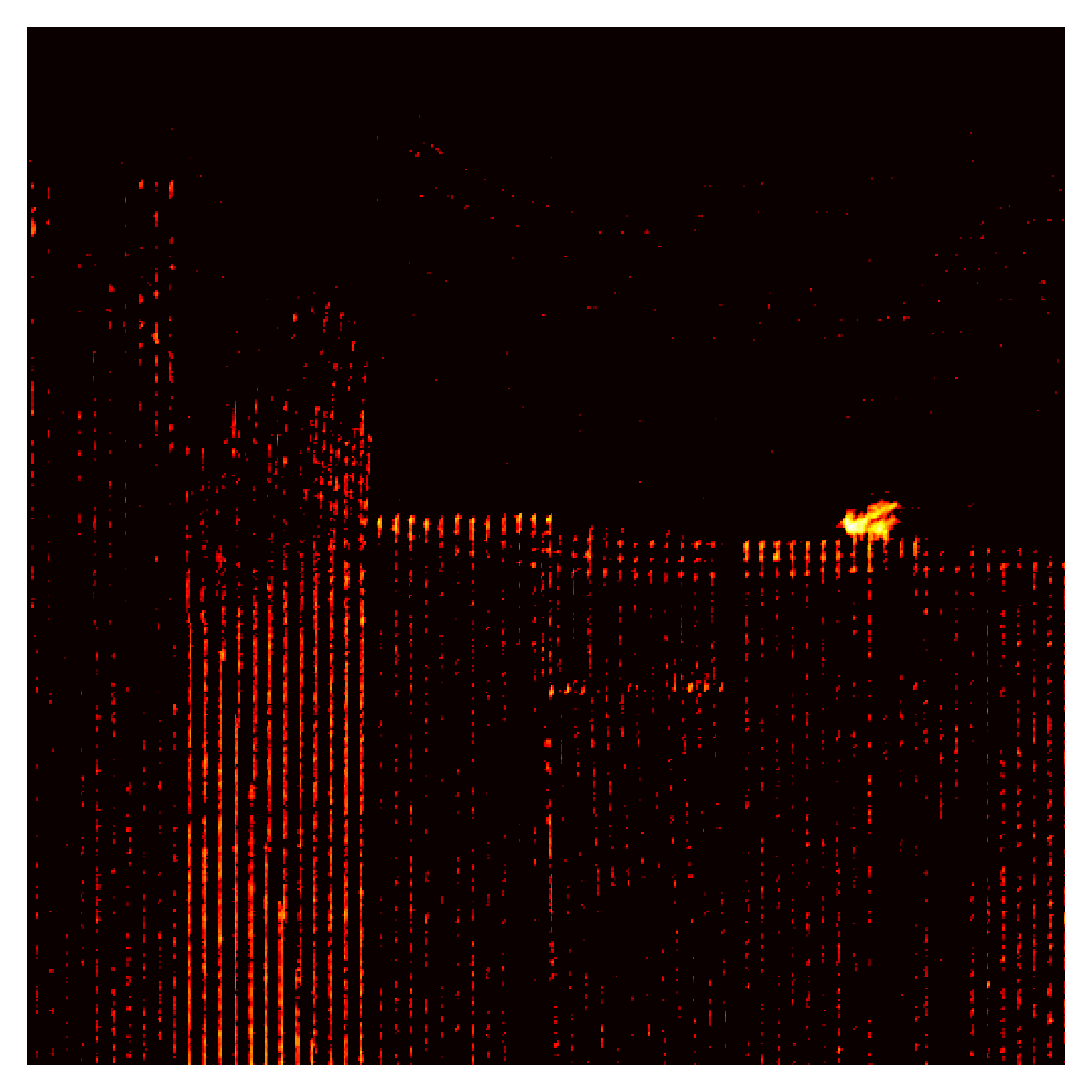}
\includegraphics[width=0.4\linewidth]{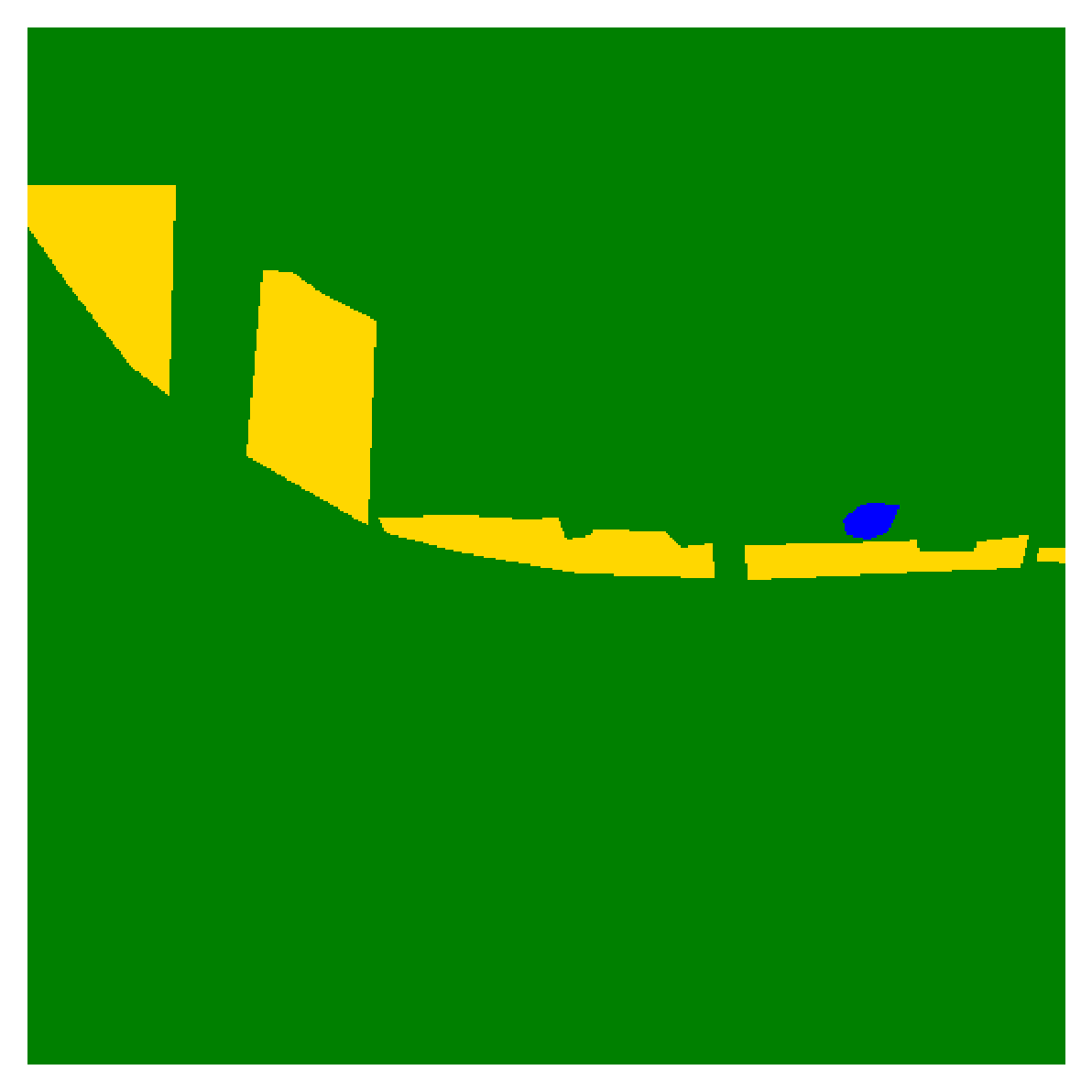}
\caption{An example of an NRGB image and its Vegetation Index (VI) and ground-truth labels.
{\bf Top-left:} Input RGB channels; {\bf Top-right:} Input near-infra red (NIR) channel;
{\bf Bottom-left: } Vegetation Condition Index (VCI)\cite{kogan1995application} calculated based on RGB and NIR channel; {\bf Bottom-right: } Ground-truth labels, where \textcolor{Dandelion}{yellow} denotes {\it Double Plant} and \textcolor{blue}{blue} denotes the {\it Weed Cluster}.
VCI is able to pick up both {\it Weed Cluster} (a cluster of very high VI values) and {\it Double Plant} (lanes of different VI values compared to the background crops).
}
\label{fig:example}

\end{figure}

To leverage the information of multiple distinct bands in the images, researchers in the last several decades have focused on developing different algorithms and metrics to perform the land segmentation\cite{gitelson2004wide, grace2007can}. As discussed in the literature review in Section~\ref{sec:relatedwork}, the design of Vegetation Index (VI) has been essential for studying land cover segmentation\cite{jordan1969derivation,xue2017significant,rouse1974monitoring, zhang1996approach}. The key idea of VI is to assess the vegetation of a region based on the reflectances from multiple bands, including the Near-Infrared band and other thermal bands, and hence ultimately approximate the region's land cover segments. Nevertheless, in the context of deep learning, we have yet to investigate how to leverage the domain knowledge of VI while making use of models learned or transferred from non-agriculture data to segment the land accurately. 

To tackle this question, we describe a general form of VI that serves as an additional input channel for image segmentation. Such a general form of VI covers many specific VI variants in existing studies\cite{rouse1974monitoring, zhang1996approach, huete2002overview, xiaoqin2015extraction}, which motivate us to develop a generalized learnable VI block that fuses the VIs and images in a convolution fashion. Based on the fused input, we also propose a new additive group normalization, a natural generalization of the instance normalization and layer normalization, because the VI channel and RGB channels can be considered as different groups.

Our work contributes to the research of agriculture land cover segmentation in three ways. Firstly, we systematically compare the vegetation indices that primarily depend on the Near-Infrared, red, green, and blue channels. We highlight the key idea of calculating VIs and disclose the connections among them.
Secondly, we propose  a model-agnostic module named General Vegetation Index (GVI) that captures many existing VI features. This module particularly fits convolutional neural networks, even for the pretrained models very well, because it doesn't need to change model structures too much. Thirdly, we introduce the additive group normalization (AGN) that helps to fine tune models smoothly when GVI is introduced to a pretrained model. With these components in place, we modified a model based on DeepLabV3\cite{chen2017rethinking} and ran experiments on land segmentation in agriculture. With careful evaluations, we achieved an mIoU of 46.89\% which exceeds the performance of the baseline model by about 2 percent.

\section{Related Work}\label{sec:relatedwork}
{\bf Vegetation Index.} Vegetation Indices (VIs) are simple and effective metrics that have been widely used to provide quantitative evaluations of vegetation growth\cite{xue2017significant}. Since the light spectrum changes with plant type, water content within tissues and so on\cite{chang2016review, zhang2012application}, the electromagnetic waves reflected from canopies can be captured by passive sensors. Such characteristics of the spectrum can provide extremely useful insights for applications in environmental and agricultural monitoring, biodiversity conservation, yield estimation, and other related fields\cite{mulla2013twenty}. Because the land vegetation highly correlates with the land cover reflectance, researchers have built more than 60 VIs in the last four decades with mainly the following light spectra: (i) the ultraviolet region (UV, 10-380 nm); the visible spectra, which consists blue (B, 450-495 nm), green (G, 495-570 nm) and red (R, 620-750 nm); (iii) the near and mid-infrared band (NIR, 850-1700 nm)\cite{rahim2016applied, cruden2012absolute}. Such VIs are validated through direct or indirect correlations with the vegetation characteristics of interest measured {\it in situ}, such as vegetation cover, biomass, growth, and vigor assessment\cite{xue2017significant, haxeltine1996general}.

To our best knowledge, the first VI, i.e., the Ratio Vegetation Index (RVI), was proposed by Jordan\cite{jordan1969derivation} in 1969. RVI was developed with the principle that leaves absorb relatively more red than infrared light. Widely used at high-density vegetation coverage regions, RVI is sensitive to atmospheric effects and noisy when vegetation cover is sparse (less than 50\%)\cite{grace2007can}. The Perpendicular Vegetation Index (PVI)\cite{richardson1977distinguishing} and the Normalized Difference Vegetation Index (NDVI)\cite{rouse1974monitoring} followed the same principle but to normalize the output, having a sensitive response even for a low vegetation coverage. To eliminate the effects of atmospheric aerosols and ozone, Kaufman and Tanre \cite{kaufman1992atmospherically} proposed the Atmospherically Resistant Vegetation Index (ARVI) in 1992, and Zhang \etal~\cite{zhang1996approach} improved the ARVI by eliminating its dependency to a 5S atmospheric transport model\cite{tanre1990technical}. Another direction was to improve VI's robustness against different soil backgrounds\cite{richardson1977distinguishing}. The Soil-Adjusted Vegetation Index (SAVI)\cite{huete1988huete} and modified SAVI (MSAVI)\cite{qi1994modified, chen1996evaluation} turned out to be much less sensitive than the RVI to changes in the background. Based on ARVI and SAVI, Liu and Huete introduced a feedback mechanism by using a parameter to simultaneously correct soil and atmospheric effects, which they called the Enhanced Vegetation Index (EVI)\cite{liu1995feedback}. With the recent progress in remote sensing (increasing number of bands and narrower bandwidth)\cite{honkavaara2013processing}, more VIs are being built to capture not only the biomass distribution and classification, but also chlorophyll content (Chlorophyll Absorption Ratio Index(CARI))\cite{kim1994use}, plant water stress (Crop Water Stress Index (CWSI))\cite{idso1981normalizing}, and light use efficiency (Photochemical Reflectance Index (PRI))\cite{ruimy1999comparing, haxeltine1996general}. With these aforementioned studies, a summary of VIs that derives from NIR-Red-Green-Blue (NRGB) images can be found in Table \ref{tab:vi}. Although we refer to a full literature review of VIs in \cite{bannari1995review} and \cite{xue2017significant}, many VIs share similar form that motivates us to find a generalized formula to capture the essence of VIs.

{\bf Remote Sensing with Transfer Learning and Data Fusion.} As an emerging interdisciplinary field, remote sensing (on both aerial photographs and satellite images) with deep learning has experienced quite a few benchmark datasets that have been released in recent years, such as FMOW\cite{christie2018functional}, SAT-4/6\cite{basu2015deepsat}, EuroSat\cite{helber2019eurosat}, DeepGlobe 2018\cite{demir2018deepglobe}, Agriculture-Vision \cite{chiu2020agriculture} and so on. Most of those datasets come with more than the visible band (i.e., RGB), including near and mid-infrared band (NIR) and sometimes shortwave red (SW). The different input structure, together with the context switch from a human-eye dataset (such as ImageNet\cite{russakovsky2015imagenet}) to a bird's-eye dataset, makes Transferring Learning less straightforward. Penatti \etal \cite{penatti2015deep} systematically compared ImageNet pretrained CNN with other descriptors (feature extractors, e.g. BIC) and found it achieve comparable but not the best performance in detecting coffee scenes. Xie \etal \cite{xie2016transfer} has shown simply adopting the ImageNet pretrained model while discarding the extra information does not achieve the best result in predicting the Poverty Level. Zhou \etal \cite{zhou2018d} has also observed the similar phenomena in their Road Extraction task. In addition, a two-stage fine-tuning process is proposed in \cite{zhou2018satellite}, where an ImageNet pretrained network is further fine-tuned on a large satellite image dataset with the first several layers frozen. An alternative direction in exploring the large-scaled but not well-labeled data is to construct satellite-image-specified geo-embedding through weakly supervised learning\cite{uzkent2019learning}, or unsupervised learning with Triplet Loss\cite{jean2019tile2vec}. These aforementioned steps motivate us to use a pretrained model based on ImageNet, which has been demonstrated to have a good performance empirically in transfer learning.

In~\cite{sidek2012review}, Sidek and Quadri defined data fusion as ``dealing with the synergistic combination of information made available by different measurement sensors, information sources, and decision-makers.'' Studies in the deep learning community have also proposed data fusion approaches that are specific to satellite images at a different level in practice. For example, \cite{castagno2018roof} concatenates LiDAR and RGB to predict roof shape better. In DeepSat \cite{basu2015deepsat}, Basu \etal achieves the state of the art performance on SAT-4 and SAT-6 land cover classification problems by incorporating NDVI\cite{rouse1974monitoring}, EVI\cite{huete2002overview} and ARVI\cite{zhang1996approach} as additional input channels. A recent study\cite{correa2015vhr} proposed a novel approach to select and combine the most similar channels using images from different timestamps. Apart from the multi-channel data fusion, fusions at multi-source \cite{schmitt2016data, gao2006blending} and multi-temporal\cite{benedetti2018m3fusion} levels have also shown their empirical value. Such an idea of fusing the multiple input channels also inspired our design of the fusion module of General Vegetation Index.

{\bf Multi-spectral Image Data Fusion.}
Multi-spectral image data fusion is also widely used in robotics, medical diagnoses, 3D inspection, etc\cite{liggins2017handbook}. {\it Color related techniques}  represent color in different spaces. The Intensity-Hue-Saturation (IHS fusion)\cite{harrison1989introduction} transforms three channels of the data into the IHS color space, which separates the color aspects in its average brightness (intensity). The values in IHS space correspond to the surface roughness, its dominant wavelength contribution (hue), and its purity (saturation) \cite{gillespie1986color, carper1990use}. Then, one of the components is replaced by a fourth channel that needs to be integrated. {\it Statistical/numerical methods} introduce a mathematical combination of image channels. The Brovey algorithm\cite{ranchin2000fusion} calculates the ratio of each image band by summing up the chosen bands, followed by multiplying with the high-resolution image.

In addition to concatenating multi-spectral channels, several deep learning architectures were proposed for multi-spectral images. \cite{piao2019new} pretrains a Siamese Convolution Network to generate a weighted map for infrared and RGB channels in the inference time. Li \etal \cite{li2018infrared} first decomposes the source images into base background and detail content and then applies a weighted average on the background while using a deep learning network to extract multi-layer features for detail content.

These studies gave an initial attempt to tackle the image classification problem using multiple spectral inputs in deep learning models. But we have yet to investigate how the multi-spectral image can be translated into the VI-related input in the context of agriculture segmentation.

\section{Proposed Method}
\subsection{Overview}
In general, our approach hinges on fusing Vegetation Index with raw images. We first introduce using well-known VIs as another input channel, and then we generalize the idea of VI to a fully learnable data fusion module. Last but not least, we propose an Additive Group Normalization (AGN) to handle the warm-start with a pretrained model. We describe the technical details in the following subsections.
\subsection{Vegetation Index for Neural Nets}
\begin{figure}[!hbpt]
\centering
\includegraphics[width=0.91\linewidth]{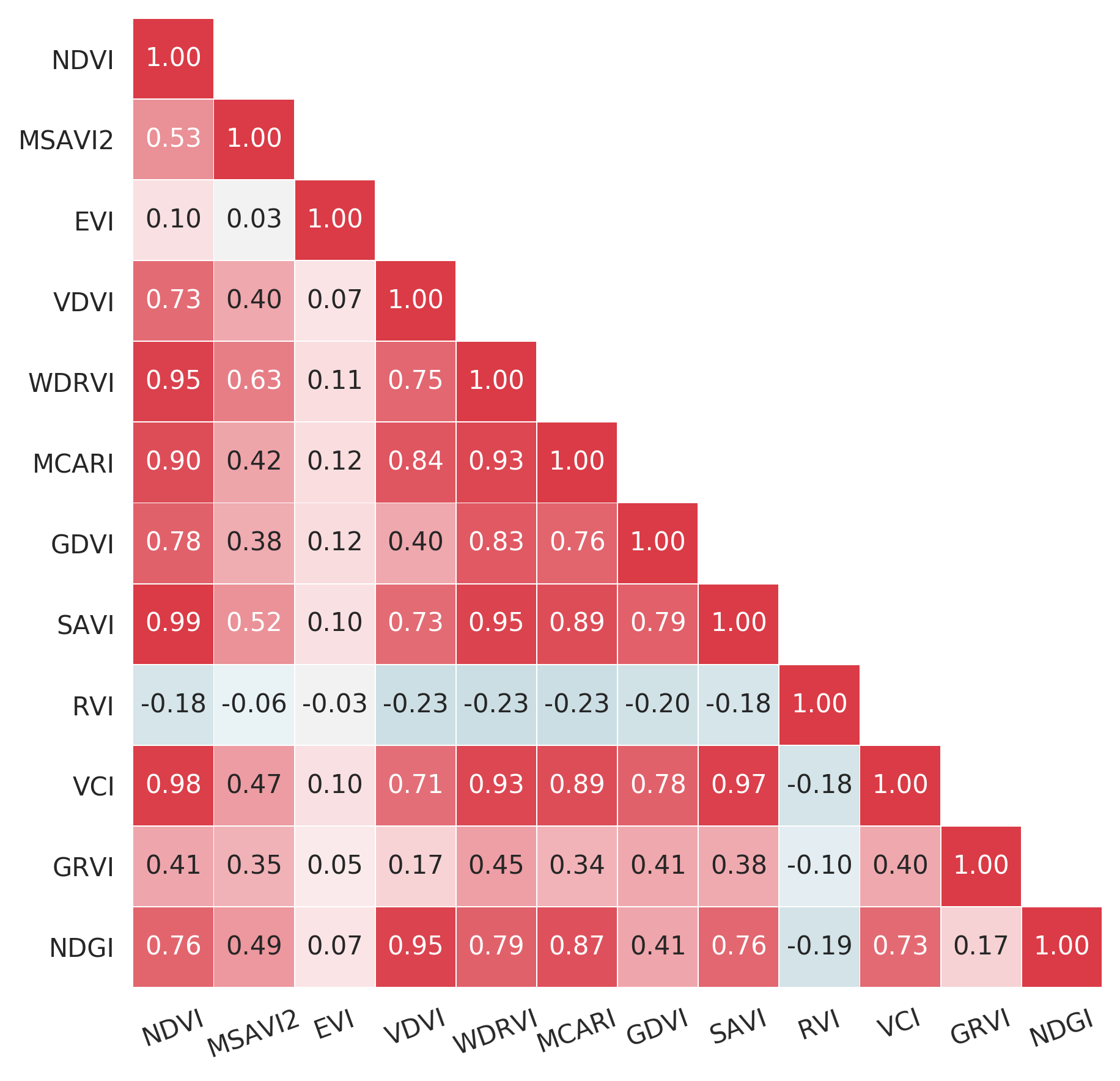}
\caption{Pair-wise correlation coefficients of all 12 available vegetation indices.}
\label{fig:vi_cm}
\end{figure}

According to \cite{xue2017significant}, during the practice of remote sensing, more than 60 VIs have been developed in the last four decades. However, not all VIs are derived from NIR and RGB channels, few of which generalize across datasets without tuning their sensitive parameters manually. For example, the Perpendicular Vegetation Index (PVI) \cite{richardson1977distinguishing} is defined as follows:
\begin{equation}
\text{PVI} = \sqrt{\left(\rho_{\text{soil}} - \rho_{\text{veg}}\right)^2_{\text{R}}  -\left(\rho_{\text{soil}} - \rho_{\text{veg}}\right)^2_{\text{NIR}}} \quad,
\end{equation}
where $\rho_{\text{soil}}$ is the soil reflectance and  $\rho_{\text{veg}}$ is the vegetation reflectivity. However, PVI is sensitive to soil brightness and reflectivity, especially in the case of low vegetation coverage, and needs to be re-calibrated for this effect\cite{kaufman1992atmospherically}. Such sensitivity introduces semantic difficulty as we try to feed the VI into the neural network as another input channel.  There are also VIs designed for a specific dataset in the first place. On top of the Landsat Multispectral Scanner (MSS), Landsat Thematic Mapper (TM) and Landsat 7 Enhanced Thematic Mapper (ETM) data, Cruden \etal~\cite{cruden2012absolute} applied a Tasseled Cap Transformation and came up with the empirical coefficients for Green Vegetation Index (GVI) as:
\begin{align}
\notag
\text{GVI} &= -0.290\text{MSS}_4 - 0.562\text{MSS}_5 + 0.600\text{MSS}_6  \\
&+ 0.49 \text{MSS}_7 ,
\end{align}
where $\text{MSS}_i$ denotes the $i$th band of Landsat MSS. Landsat TM and Landsat 7 ETM are not usually available for satellite and aerial imagery outside this product family. This Green VI is composed by a linear combination of the multi-channel input, which shares a similar concept among many other VIs. To better understand the popular format of different indices, we summarized some representative VIs, shown in Table~\ref{tab:vi}, that are derived from NIR-Red-Green-Blue (NRGB) images, together with their definitions and value ranges. Based on the definitions, we calculate the pixel-wise correlation matrix for all 12 VIs (Figure \ref{fig:vi_cm}). The correlation coefficients are calculated at the pixel level using all data released for training. For SAVI, we choose $L$ to be 0.5.
\begin{table*}[ht]
\centering
\begin{small}
\begin{tabular}{l|c|c}
\toprule
Index & Definition & Meaningful Range\\
\midrule 
NDVI$^*$  \cite{rouse1974monitoring} &  $\displaystyle\frac{\text{NIR} - \text{R}}{\text{NIR} + \text{R}}$ & $[0, 1]$\\[10pt]
IAVI$^*$$^\dagger$ \cite{zhang1996approach} & $\displaystyle\frac{\text{NIR} - (\text{R} - \gamma (\text{B} - \text{R}))}{\text{NIR} + (\text{R} - \gamma (\text{B} - \text{R}))}$, $\gamma \in (0.65, 1.12)$ & $[-1, 1]$\\[10pt]
MSAVI2$^*$ \cite{richardson1977distinguishing} & 0.5 $\displaystyle\left((2\text{NIR} + 1) - \sqrt{(2\text{NIR}+1)^2 -8(NIR-R)}\right)$ & $[0, 1]$\\[10pt]
EVI$^*$ \cite{huete2002overview} & $\displaystyle2.5*\frac{\text{NIR} - \text{R}}{ \text{NIR} + 6\text{R} - 7.5\text{B} + 1}$ & $(-\infty, \infty)$\\[10pt]
VDVI$^*$ \cite{xiaoqin2015extraction}& $\displaystyle2*\frac{2\text{G} - \text{R} -\text{B}}{2\text{G} + \text{R} + \text{B}}$ & $[-1, 1]$\\[10pt]
WDRVI$^*$ \cite{gitelson2004wide} & $\displaystyle\frac{0.2\text{NIR} - \text{R}}{0.2\text{NIR} + \text{R}}$ & $[-1, 1]$\\[10pt]
MCARI \cite{daughtry2000estimating} & $\displaystyle\frac{1.5*(2.5*(\text{NIR} - \text{R}) -1.3 *(\text{NIR}-\text{G})}{\sqrt{(2\text{NIR} + 1)^2 - (6\text{NIR} -5\text{R})-0.5}}$  & $(-1.6, 4.88)$ \\[10pt]
GDVI \cite{sripada2005aerial} &$\displaystyle\text{NIR} - \text{G}$ & $[-1, 1]$\\[10pt]
SAVI$^*$$^\dagger$  \cite{huete1988huete} & $\displaystyle (1+L)*\frac{\text{NIR} - \text{R}}{ \text{NIR} + \text{R} + L}, \, L\in\{0, 0.5, 1\}$ & $[0,1]$\\[10pt]
RVI$^*$  \cite{pearson1972remote} & $\displaystyle \frac{\text{R}}{\text{NIR}}$ & $[0, \infty)$\\[10pt]
VCI \cite{kogan1995application} & $\displaystyle\frac{\text{NDVI}-\text{NDVI}_{\text{min}}}{\text{NDVI}_{\text{max}}+\text{NDVI}_{\text{min}}}$ & $[0,1]$\\[10pt]
GRVI$^*$ \cite{sripada2005aerial} & $\displaystyle \frac{\text{NIR}}{\text{G}}$ & $[0, \infty)$\\[10pt]
NDGI$^*$  \cite{baret1991potentials} &  $\displaystyle\frac{\text{G} - \text{R}}{\text{G} + \text{R}}$ & $[-1, 1]$\\[10pt]
\bottomrule
\end{tabular}
\vspace{0.05in}
\caption{Summary of vegetation indices that are derived from NIR-Red-Green-Blue (NRGB) images}
$^*$: These vegetation indices share the general format as equation \ref{eq:vi} \\
$^\dagger$: Parameters need to be calibrated and this VI cannot be fed into the neural network directly
\label{tab:vi}
\end{small}
\end{table*}
Except for certain pairs (such as NDVI v.s. SAVI), the correlation between different VIs are within the range of $(-0.2, 0.9)$. We include all 12 VIs as extra input channels in our experiments when leveraging the information from existing vegetation indices.
\subsection{Learnable Vegetation Index}
\begin{figure}[!hbpt]
\centering
\includegraphics[width=0.95\linewidth]{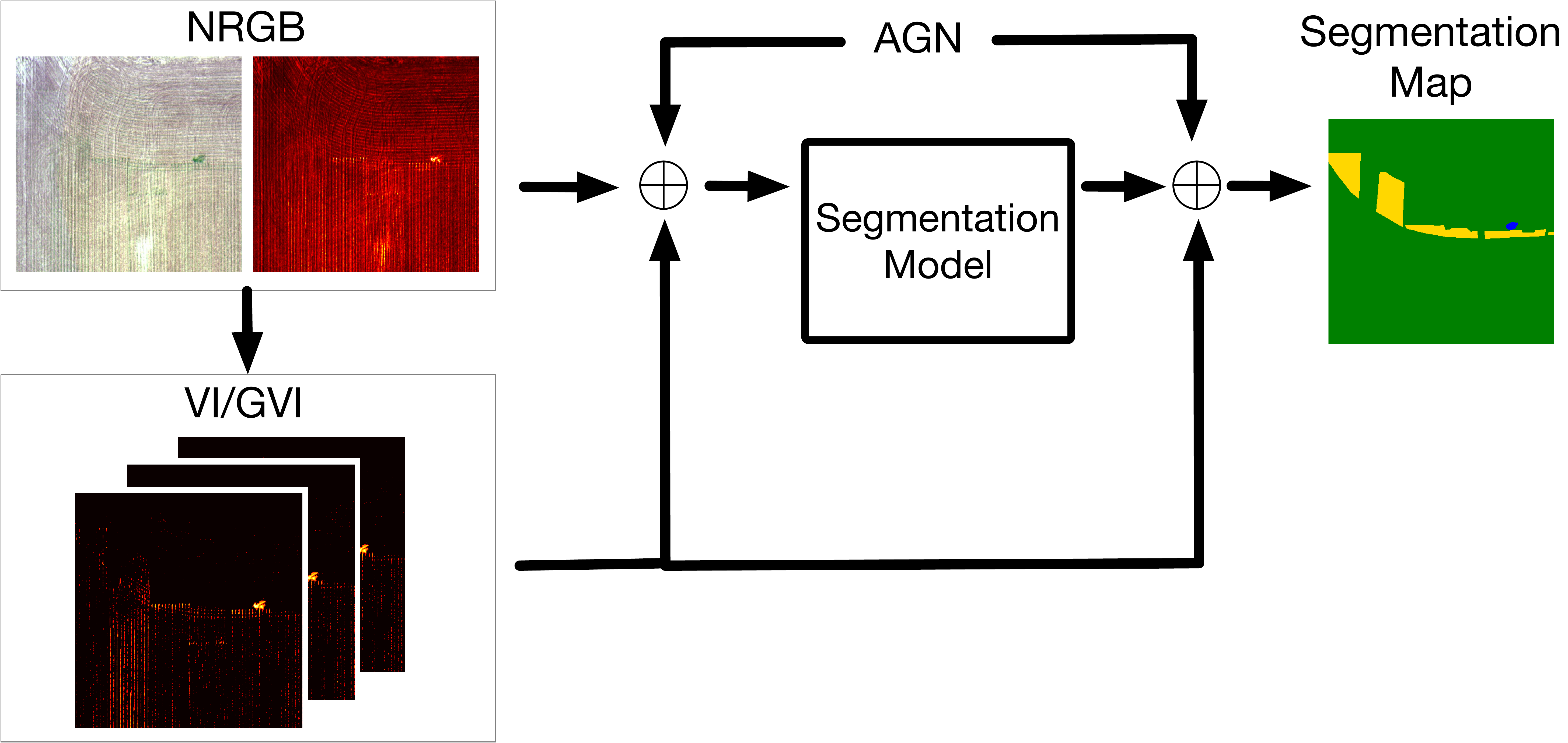}
\caption{Our data fusion module is model-agnostic. The VI or GVI input channel is compatible to any segmentation model trained on NRGB image. Additive Group Normalization (AGN) is applied in the Near-Infrared channel with a linear combination of the batch normalization. }
\label{fig:model}
\end{figure}

Some high correlations between VIs stem from not only the fundamental vegetation status, but also the empirical function that researchers have introduced. We notice that 9 out of the 13 VIs from Table \ref{tab:vi} share the following general form:
\begin{equation} \label{eq:vi}
\text{VI} = \frac{\alpha_0+\alpha_{\text{R}} \text{R} + \alpha_{\text{G}} \text{G}+ \alpha_{\text{B}} \text{B}+ \alpha_{\text{NIR}}\text{NIR}}{\beta_0+\beta_{\text{R}} \text{R} + \beta_{\text{G}} \text{G}+ \beta_{\text{B}} \text{B}+ \beta_{\text{NIR}}\text{NIR}},
\end{equation}
where $\alpha_c, \beta_c, \, \forall c\in \{\text{R}, \text{G}, \text{B}, \text{NIR}\}$ are parameters to be determined, and could be learnable in deep learning models. As suggested in \cite{kogan1995application}, we can normalize the response (output) by nearby regions to supress outliers. By extending the pixel-wise operation to each neighborhood of image channels, we introduce a learnable layer of Generalized Vegetation Index  (GVI):
\begin{equation} \label{eq:gvi}
GVI(\mathbf{x}, \mathbf{\alpha}, \mathbf{\beta}) = \frac{\mathbf{x} \circledast \mathbf{\alpha}}{\mathbf{x} \circledast \mathbf{\beta}},
\end{equation}
where $\circledast$ denotes the convolution operation, $\mathbf{x}$ is our NRGB inputs, and $\mathbf{\alpha}, \mathbf{\beta}$ are the learnable weights. In practice, we clip both the numerator and denominator to avoid numerical issues. Depending on the output channels, this layer has the capacity to express a variant number of VIs when learned. An illustrative example can be found in Figure~\ref{fig:model}.
\subsection{Additive Group Normalization Index}
\begin{figure}[!hpbt]
\centering
\includegraphics[width=0.75\linewidth]{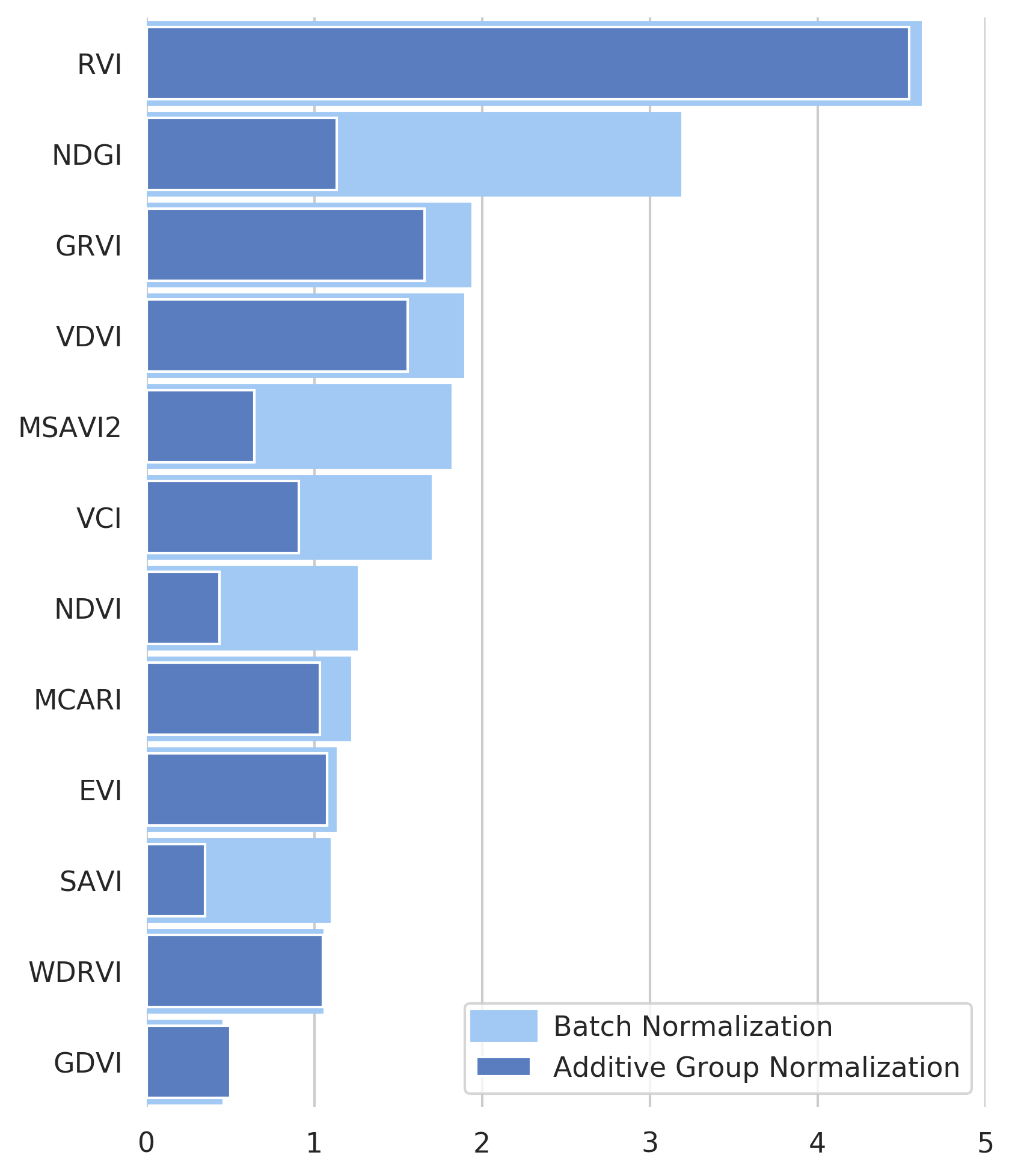}
\caption{Mean L1 error over standard deviation (\%).
At each pixel, we trained a two-layer, fully-connected neural network with the NRGB channels to fit the Vegetation Indices using a batch size of 16. We plot the relative error for each vegetation index, i.e., the L1 error over the mean standard deviation in percentage. The additive group normalization fits almost all VIs better compared to batch normalization. }
\label{fig:vi_err}
\end{figure}

In contrast to explicitly normalizing the constructed indices using a ratio (e.g., in Equation \eqref{eq:gvi}), we could normalize the value using nearby regions and channels, as we saw in VCI\cite{kogan1995application}. Fortunately, the deep learning community has already developed the counterpart approaches, such as the Batch Normalization (BN)\cite{ioffe2015batch}, Layer Normalization (LN)\cite{ba2016layer}, Instance Normalization (IN)\cite{ulyanov2016instance} and Group Normalization (GN)\cite{wu2018group}. However, we found that the neural network, even equipped with the most widely used BN, has an internal difficulty in fitting existing VIs, as shown in Figure \ref{fig:vi_err}.

%
The relative high errors in prediction indicate that BN is not able to captured channel-normalized features, while VIs are usually normalizing the inputs across the spectrum. Motivated by such observations, we introduce the Additive Group Normalization, which combine the BN and GM together in an additive fashion. Unlike BN, which normalizes each channel of features using the mean and variance computed over the mini-batch, GN splits channels into groups and uses the within-group mean and variance to normalize the particular group:
\begin{align}\notag
\hat{x}_{nchw}^{GN} &= \frac{x_{nchw} - \mu_{nc}^{(GN)}}{\sqrt{\sigma_{nc}^{2(GN)} +\epsilon}}\\ \notag
\mu_{nc}^{(GN)} &= \frac{1}{HWG}\sum_{c\in G_c}\sum_{H}\sum_{W} x_{nchw}\\
\sigma_{nc}^{2(GN)} &= \frac{1}{HWG}\sum_{c\in G_c}\sum_{H}\sum_{W} \left( x_{nchw} -\mu_{nc}^{(GN)} \right),
\end{align}
where $G$ is the number of groups, $G_c$ is the group assignment of channel $c$, and $\hat{\mathbf{x}}^{(GN)} = \{\hat{x}_{nchw}^{(GN)}\}$ is the GN response. Depending on the number of groups, such a normalization can be reduced to either Instance Normalization ($G=C$) or Layer Normalization ($G=1$).
\begin{figure*}[!hbpt]
\centering
\includegraphics[width=0.7\linewidth]{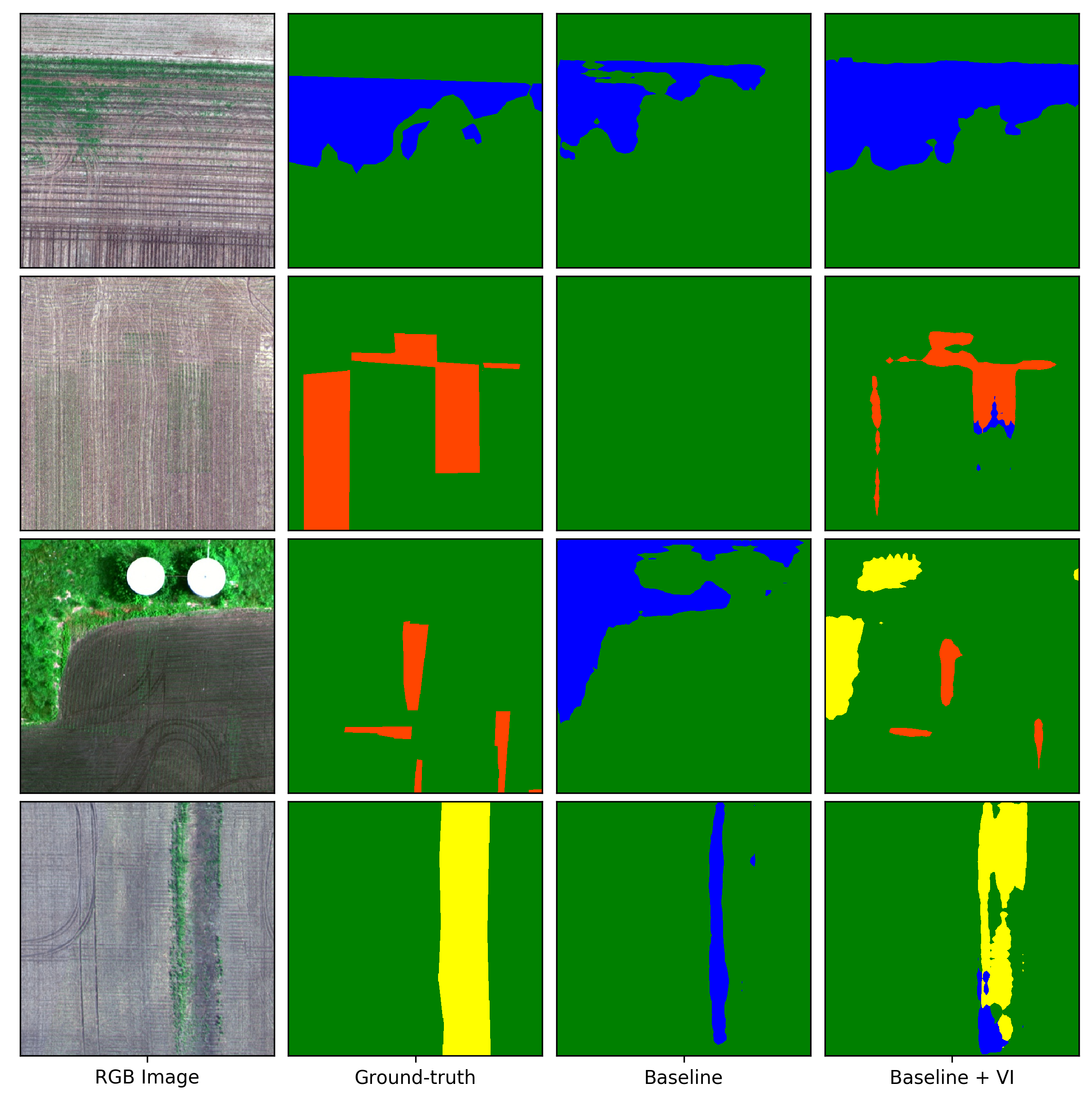}
\caption{Examples of segmentation results.
We include four examples (rows) of their RGB input, ground-truth labels, predictions of the baseline model, and predictions of the baseline model with VIs inputs (columns).
Segmentation labels: \textcolor{ForestGreen}{Green} for {\it background}, \textcolor{blue}{Blue} for {\it Weed Cluster}, \textcolor{red}{Red} for {\it Double Plant} and \textcolor{Dandelion}{Yellow} for {\it Waterway}. Including VIs helps the model perform better in vegetation-related classes (e.g. {\it Weed Cluster}) as well as non vegetation classes (e.g. {\it Waterway}).
}
\label{fig:seg}
\end{figure*}

Inspired by the adaptive Instance-Batch Normalization \cite{nam2018batch}, we designed our Additive Group Normalization (AGN) as follows:
\begin{equation}
\hat{\mathbf{x}}^{(AGN)} = \sigma(\rho) \cdot \hat{\mathbf{x}}^{(GN)} + \hat{\mathbf{x}}^{(BN)} ,
\end{equation}
where $\rho$ is a learnable parameter controlling the contribution of Group Normalization in each layer and $\hat{\mathbf{x}}^{(AGN)} \in \mathcal{R}^{N\times C \times H \times W}$ is the response of AGN. This normalization does not introduce extra parameters (except for the running mean and standard deviation) but leverages the existing capacity of the underlying network.

When $\rho$ is a large negative number, the term $\hat{\mathbf{x}}^{(G)}$ gets a negligible weight and $ \hat{\mathbf{x}}^{(AGN)} \approx \hat{\mathbf{x}}^{(B)}$. This property makes fine-tuning of experiments much smoother on a pretrained model with an architecture of Batch Normalization: To control the ``ramping up'' of Group Normalization, we initialize $\rho$ with a negative number, e.g., $-10$, and the model weights are updated gradually. We show experimental results for both training from scratch and fine-tuning in Section~\ref{sec:expriments}.
\section{Experiments}\label{sec:expriments}
\subsection{Architecture Setup}
We use EfficientNet-B0 / EfficientNet-B2\cite{tan2019efficientnet} as our base encoder in the DeepLabV3\cite{chen2017rethinking} framework. They are parameter-efficient networks that achieve the same performance of ResNet-50 / ResNet-101 respectively with a much lower number of parameters.\cite{he2016deep}

\begin{table*}[!hpbt]
\centering
\begin{small}
\begin{tabular}{c c | c | c c c c c c c}
\toprule
Architecture & Method & mIoU (\%) & Background & \makecell{Cloud \\ Shadow} & \makecell{Double \\ Plant} &\makecell{Planter \\ Skip} & \makecell{Standing \\ Water} &Waterway &\makecell{Weed\\ Cluster}\\
\midrule
DeepLabV3 & Baseline &44.92    &78.84    &40.59    &33.14    &0.74    &51.03    &60.64    &49.48\\
& Baseline + VI &46.04    &78.75    &41.17    &33.66    &0.46    &56.67    &62.06    &49.50\\
& Baseline + GVI &46.05    &{\bf 79.81}    &34.58    &{\bf 35.24}    &0.83    &{\bf 58.08}    &63.47    &50.32\\
& AGN &{\bf 46.87}    &79.28    &{\bf 41.22}    &34.56    &{\bf 1.05}    &57.14    &{\bf 63.53}    &{\bf 51.28}\\
\bottomrule
\end{tabular}
\end{small}
\vspace{0.1in}
\caption{mIoUs and class IoUs of baseline models, baseline models with Vegetation Index as additional models and our proposed generalized vegetation index model.}
\label{tab:result}
\end{table*}
\subsection{Training Details}
We used backbone models pretrained on ImageNet in all our experiments. During initialization, we copied the pretrained weights for the red channel filter to the one for the NIR channel in the first layer. We trained each model for 80 epochs with a batch size of 64 on eight GeForce GTX TITAN X GPUs. Unless specified, we used a combination of Focal Loss\cite{lin2017focal} and Dice Loss\cite{sudre2017generalised} with weights 0.75 and 0.25 respectively. We did not weigh classes differently, albeit the dataset is unbalanced. We also masked all the pixels that are either not valid or not within the region of the farmland. We use the Adam optimizer \cite{kingma2014adam} with a base learning rate of 0.01 and a weight decay of $5 \times 10^{-4}$. During the training, we monitored the validation loss and stopped the experiments if the loss didn't decrease within ten epochs. Once a model was trained, we fine-tuned it with VI, GVI, or AGN modules. We adopted the cosine annealing strategy \cite{loshchilov2016sgdr}, with the learning rate ranges from $0.0001$ to $0.01$ and a cycle length of 10 epochs.  For a fair comparison, we also fine-tuned our baseline model in this stage.

\subsection{Dataset and Evaluation Metric}
We evaluated our approach on Agriculture-Vision\cite{chiu2020agriculture} with mean Intersection-over-Union(IOU) across classes. Since our annotations may overlap while we modeled the segmentation as a multi-class classification problem pixel-wisely, we also described the mean IOU calculation as follows.\\
{\bf Agriculture-Vision.} Agriculture-Vision is an aerial image dataset that contains 21,061 farmland images captured throughout 2019 across the US. Each image is of size 512 $\times$ 512 and with four color channels, namely, RGB and Near Infrared (NIR). By the time the experiments are done, the labels for the test set have not been released yet, so we used the veriﬁcation set to test the trained model. \\
{\bf IOU with overlapped annotations.} We followed the protocol from the data challenge organizer to accommodate the evaluation for overlapped annotations. For pixels with multiple labels, a prediction of either label was counted as a correct pixel classification for that label, and a prediction that did not contain any ground truth labels was counted as an incorrect classification for all ground truth labels.
\subsection{Results}
Table \ref{tab:result} presents the validation results of the baseline model, together with several proposed methods to leverage the information from the NIR band. The average mIoU of the baseline model yeilds 44.92\% accuracy. When plugging the GVI module into our model, we achieve 46.05\% accuracy on mIoU, and highest accuracy in some categories such as \emph{background}, \emph{double plant}, and \emph{standing water}. Moreover, when we use the Additive Group Normalization (AGN), the model performs the best in terms of mIoU at the accuracy of 46.87\%.  Our model consistently outperforms a) only using the NIR bands without extra information from vegetation indices; b) adding vegetation indices directly as inputs. And we saw gains in vegetation-related classes (e.g., Weed Cluster) as well as non-vegetation classes (e.g., Waterway). We include some examples in Figure \ref{fig:seg}.
\vspace{-0.09in}
\section{Conclusion}
\vspace{-0.06in}
In this work, we introduced the \emph{General Vegetation Index} that enhanced the power of neural networks in agriculture and highlighted the connection between this GVI and other existing VIs. When starting from a pretrained model with minimal modifications, our proposed GVI and additive group normalization can achieve, and in some cases, exceed state-of-the-art performances. Our best result of mIOU is about 2\% better than the baseline model. In addition, our method doesn't require sophisticated network architecture with the increase of model parameters. Such a result is a promising step forward when incorporating VI related information with multi-band images for segmentation tasks in agriculture.

While our approach sheds a promising light on segmenting lands in agriculture, we believe several potential directions could be valuable for future work. Firstly, how the model architecture can affect the result is still open for exploring. It is not clear if the segmentation results are sensitive to different models with VI inputs. Secondly, we would like to incorporate some additional training techniques, e.g., virtual adversarial training, which is orthogonal to our data fusion approach to improve the model performance further.  Lastly, the ability to generalize our method on a larger scale dataset remains open to investigate.

{
\small
\bibliographystyle{unsrtnat}
\bibliography{egbib}

\begin{thebibliography}{71}
\providecommand{\natexlab}[1]{#1}
\providecommand{\url}[1]{\texttt{#1}}
\expandafter\ifx\csname urlstyle\endcsname\relax
  \providecommand{\doi}[1]{doi: #1}\else
  \providecommand{\doi}{doi: \begingroup \urlstyle{rm}\Url}\fi

\bibitem[Irvin et~al.(2019)Irvin, Rajpurkar, Ko, Yu, Ciurea-Ilcus, Chute,
  Marklund, Haghgoo, Ball, Shpanskaya, et~al.]{irvin2019chexpert}
Jeremy Irvin, Pranav Rajpurkar, Michael Ko, Yifan Yu, Silviana Ciurea-Ilcus,
  Chris Chute, Henrik Marklund, Behzad Haghgoo, Robyn Ball, Katie Shpanskaya,
  et~al.
\newblock Chexpert: A large chest radiograph dataset with uncertainty labels
  and expert comparison.
\newblock In \emph{Proceedings of the AAAI Conference on Artificial
  Intelligence}, volume~33, pages 590--597, 2019.

\bibitem[Oshri et~al.(2018)Oshri, Hu, Adelson, Chen, Dupas, Weinstein, Burke,
  Lobell, and Ermon]{oshri2018infrastructure}
Barak Oshri, Annie Hu, Peter Adelson, Xiao Chen, Pascaline Dupas, Jeremy
  Weinstein, Marshall Burke, David Lobell, and Stefano Ermon.
\newblock Infrastructure quality assessment in africa using satellite imagery
  and deep learning.
\newblock In \emph{Proceedings of the 24th ACM SIGKDD International Conference
  on Knowledge Discovery \& Data Mining}, pages 616--625, 2018.

\bibitem[Karpathy et~al.(2014)Karpathy, Toderici, Shetty, Leung, Sukthankar,
  and Fei-Fei]{karpathy2014large}
Andrej Karpathy, George Toderici, Sanketh Shetty, Thomas Leung, Rahul
  Sukthankar, and Li~Fei-Fei.
\newblock Large-scale video classification with convolutional neural networks.
\newblock In \emph{Proceedings of the IEEE conference on Computer Vision and
  Pattern Recognition}, pages 1725--1732, 2014.

\bibitem[Chiu et~al.(2020)Chiu, Xu, Wei, Huang, Schwing, Brunner, Khachatrian,
  Karapetyan, Dozier, Rose, et~al.]{chiu2020agriculture}
Mang~Tik Chiu, Xingqian Xu, Yunchao Wei, Zilong Huang, Alexander Schwing,
  Robert Brunner, Hrant Khachatrian, Hovnatan Karapetyan, Ivan Dozier, Greg
  Rose, et~al.
\newblock Agriculture-vision: A large aerial image database for agricultural
  pattern analysis.
\newblock \emph{arXiv preprint arXiv:2001.01306}, 2020.

\bibitem[Kogan(1995)]{kogan1995application}
Felix~N Kogan.
\newblock Application of vegetation index and brightness temperature for
  drought detection.
\newblock \emph{Advances in space research}, 15\penalty0 (11):\penalty0
  91--100, 1995.

\bibitem[Gitelson(2004)]{gitelson2004wide}
Anatoly~A Gitelson.
\newblock Wide dynamic range vegetation index for remote quantification of
  biophysical characteristics of vegetation.
\newblock \emph{Journal of plant physiology}, 161\penalty0 (2):\penalty0
  165--173, 2004.

\bibitem[Grace et~al.(2007)Grace, Nichol, Disney, Lewis, Quaife, and
  Bowyer]{grace2007can}
J~Grace, C~Nichol, M~Disney, P~Lewis, Tristan Quaife, and P~Bowyer.
\newblock Can we measure terrestrial photosynthesis from space directly, using
  spectral reflectance and fluorescence?
\newblock \emph{Global Change Biology}, 13\penalty0 (7):\penalty0 1484--1497,
  2007.

\bibitem[Jordan(1969)]{jordan1969derivation}
Carl~F Jordan.
\newblock Derivation of leaf-area index from quality of light on the forest
  floor.
\newblock \emph{Ecology}, 50\penalty0 (4):\penalty0 663--666, 1969.

\bibitem[Xue and Su(2017)]{xue2017significant}
Jinru Xue and Baofeng Su.
\newblock Significant remote sensing vegetation indices: A review of
  developments and applications.
\newblock \emph{Journal of Sensors}, 2017, 2017.

\bibitem[Rouse et~al.(1974)Rouse, Haas, Schell, and
  Deering]{rouse1974monitoring}
JW~Rouse, RH~Haas, JA~Schell, and DW~Deering.
\newblock Monitoring vegetation systems in the great plains with erts.
\newblock \emph{NASA special publication}, 351:\penalty0 309, 1974.

\bibitem[Zhang et~al.(1996)Zhang, Rao, and Liao]{zhang1996approach}
Renhua Zhang, XN~Rao, and NK~Liao.
\newblock Approach for a vegetation index resistant to atomospheric effect.
\newblock \emph{Acta Botanica Sinica}, 38\penalty0 (1):\penalty0 53--62, 1996.

\bibitem[Huete et~al.(2002)Huete, Didan, Miura, Rodriguez, Gao, and
  Ferreira]{huete2002overview}
Alfredo Huete, Kamel Didan, Tomoaki Miura, E~Patricia Rodriguez, Xiang Gao, and
  Laerte~G Ferreira.
\newblock Overview of the radiometric and biophysical performance of the modis
  vegetation indices.
\newblock \emph{Remote sensing of environment}, 83\penalty0 (1-2):\penalty0
  195--213, 2002.

\bibitem[Xiaoqin et~al.(2015)Xiaoqin, Miaomiao, Shaoqiang, and
  Yundong]{xiaoqin2015extraction}
Wang Xiaoqin, Wang Miaomiao, Wang Shaoqiang, and Wu~Yundong.
\newblock Extraction of vegetation information from visible unmanned aerial
  vehicle images.
\newblock \emph{Transactions of the Chinese Society of Agricultural
  Engineering}, 31\penalty0 (5), 2015.

\bibitem[Chen et~al.(2017)Chen, Papandreou, Schroff, and
  Adam]{chen2017rethinking}
Liang-Chieh Chen, George Papandreou, Florian Schroff, and Hartwig Adam.
\newblock Rethinking atrous convolution for semantic image segmentation.
\newblock \emph{arXiv preprint arXiv:1706.05587}, 2017.

\bibitem[Chang et~al.(2016)Chang, Peng-Sen, and Shi-Rong]{chang2016review}
L~Chang, S~Peng-Sen, and Liu Shi-Rong.
\newblock A review of plant spectral reflectance response to water
  physiological changes.
\newblock \emph{Chinese Journal of Plant Ecology}, 40\penalty0 (1):\penalty0
  80--91, 2016.

\bibitem[Zhang and Kovacs(2012)]{zhang2012application}
Chunhua Zhang and John~M Kovacs.
\newblock The application of small unmanned aerial systems for precision
  agriculture: a review.
\newblock \emph{Precision agriculture}, 13\penalty0 (6):\penalty0 693--712,
  2012.

\bibitem[Mulla(2013)]{mulla2013twenty}
David~J Mulla.
\newblock Twenty five years of remote sensing in precision agriculture: Key
  advances and remaining knowledge gaps.
\newblock \emph{Biosystems engineering}, 114\penalty0 (4):\penalty0 358--371,
  2013.

\bibitem[Rahim et~al.(2016)Rahim, Lokman, Harun, Hornyak, Sterckx, Mohammed,
  and Dutta]{rahim2016applied}
Hazli Rafis Bin~Abdul Rahim, Muhammad Quisar~Bin Lokman, Sulaiman~Wadi Harun,
  Gabor~Louis Hornyak, Karel Sterckx, Waleed~Soliman Mohammed, and Joydeep
  Dutta.
\newblock Applied light-side coupling with optimized spiral-patterned zinc
  oxide nanorod coatings for multiple optical channel alcohol vapor sensing.
\newblock \emph{Journal of Nanophotonics}, 10\penalty0 (3):\penalty0 036009,
  2016.

\bibitem[Cruden et~al.(2012)Cruden, Prabhu, and Martinez]{cruden2012absolute}
Brett~A Cruden, Dinesh Prabhu, and Ramon Martinez.
\newblock Absolute radiation measurement in venus and mars entry conditions.
\newblock \emph{Journal of Spacecraft and Rockets}, 49\penalty0 (6):\penalty0
  1069--1079, 2012.

\bibitem[Haxeltine and Prentice(1996)]{haxeltine1996general}
Alex Haxeltine and IC~Prentice.
\newblock A general model for the light-use efficiency of primary production.
\newblock \emph{Functional Ecology}, pages 551--561, 1996.

\bibitem[Richardson and Wiegand(1977)]{richardson1977distinguishing}
Arthur~J Richardson and CL~Wiegand.
\newblock Distinguishing vegetation from soil background information.
\newblock \emph{Photogrammetric engineering and remote sensing}, 43\penalty0
  (12):\penalty0 1541--1552, 1977.

\bibitem[Kaufman and Tanre(1992)]{kaufman1992atmospherically}
Yoram~J Kaufman and Didier Tanre.
\newblock Atmospherically resistant vegetation index (arvi) for eos-modis.
\newblock \emph{IEEE transactions on Geoscience and Remote Sensing},
  30\penalty0 (2):\penalty0 261--270, 1992.

\bibitem[Tanr{\'e} et~al.(1990)Tanr{\'e}, Deroo, Duhaut, Herman, Morcrette,
  Perbos, and Deschamps]{tanre1990technical}
D~Tanr{\'e}, C~Deroo, P~Duhaut, M~Herman, JJ~Morcrette, J~Perbos, and
  PY~Deschamps.
\newblock Technical note description of a computer code to simulate the
  satellite signal in the solar spectrum: the 5s code.
\newblock \emph{International Journal of Remote Sensing}, 11\penalty0
  (4):\penalty0 659--668, 1990.

\bibitem[Huete(1988)]{huete1988huete}
Alfredo Huete.
\newblock Huete, ar a soil-adjusted vegetation index (savi). remote sensing of
  environment.
\newblock \emph{Remote sensing of environment}, 25:\penalty0 295--309, 1988.

\bibitem[Qi et~al.(1994)Qi, Chehbouni, Huete, Kerr, and
  Sorooshian]{qi1994modified}
Jiaguo Qi, Abdelghani Chehbouni, Alfredo~R Huete, Yann~H Kerr, and Soroosh
  Sorooshian.
\newblock A modified soil adjusted vegetation index.
\newblock 1994.

\bibitem[Chen(1996)]{chen1996evaluation}
Jing~M Chen.
\newblock Evaluation of vegetation indices and a modified simple ratio for
  boreal applications.
\newblock \emph{Canadian Journal of Remote Sensing}, 22\penalty0 (3):\penalty0
  229--242, 1996.

\bibitem[Liu and Huete(1995)]{liu1995feedback}
Hui~Qing Liu and Alfredo Huete.
\newblock A feedback based modification of the ndvi to minimize canopy
  background and atmospheric noise.
\newblock \emph{IEEE transactions on Geoscience and Remote Sensing},
  33\penalty0 (2):\penalty0 457--465, 1995.

\bibitem[Honkavaara et~al.(2013)Honkavaara, Saari, Kaivosoja, P{\"o}l{\"o}nen,
  Hakala, Litkey, M{\"a}kynen, and Pesonen]{honkavaara2013processing}
Eija Honkavaara, Heikki Saari, Jere Kaivosoja, Ilkka P{\"o}l{\"o}nen, Teemu
  Hakala, Paula Litkey, Jussi M{\"a}kynen, and Liisa Pesonen.
\newblock Processing and assessment of spectrometric, stereoscopic imagery
  collected using a lightweight uav spectral camera for precision agriculture.
\newblock \emph{Remote Sensing}, 5\penalty0 (10):\penalty0 5006--5039, 2013.

\bibitem[Kim et~al.(1994)Kim, Daughtry, Chappelle, McMurtrey, and
  Walthall]{kim1994use}
Moon~S Kim, CST Daughtry, EW~Chappelle, JE~McMurtrey, and CL~Walthall.
\newblock The use of high spectral resolution bands for estimating absorbed
  photosynthetically active radiation (a par).
\newblock 1994.

\bibitem[Idso et~al.(1981)Idso, Jackson, Pinter~Jr, Reginato, and
  Hatfield]{idso1981normalizing}
SB~Idso, RD~Jackson, PJ~Pinter~Jr, RJ~Reginato, and JL~Hatfield.
\newblock Normalizing the stress-degree-day parameter for environmental
  variability.
\newblock \emph{Agricultural meteorology}, 24:\penalty0 45--55, 1981.

\bibitem[Ruimy et~al.(1999)Ruimy, Kergoat, Bondeau, and
  Intercomparison]{ruimy1999comparing}
A~Ruimy, L~Kergoat, Alberte Bondeau, and ThE Participants OF ThE Potsdam
  NpP~Model Intercomparison.
\newblock Comparing global models of terrestrial net primary productivity
  (npp): Analysis of differences in light absorption and light-use efficiency.
\newblock \emph{Global Change Biology}, 5\penalty0 (S1):\penalty0 56--64, 1999.

\bibitem[Bannari et~al.(1995)Bannari, Morin, Bonn, and
  Huete]{bannari1995review}
A~Bannari, D~Morin, F~Bonn, and AR~Huete.
\newblock A review of vegetation indices.
\newblock \emph{Remote sensing reviews}, 13\penalty0 (1-2):\penalty0 95--120,
  1995.

\bibitem[Christie et~al.(2018)Christie, Fendley, Wilson, and
  Mukherjee]{christie2018functional}
Gordon Christie, Neil Fendley, James Wilson, and Ryan Mukherjee.
\newblock Functional map of the world.
\newblock In \emph{Proceedings of the IEEE Conference on Computer Vision and
  Pattern Recognition}, pages 6172--6180, 2018.

\bibitem[Basu et~al.(2015)Basu, Ganguly, Mukhopadhyay, DiBiano, Karki, and
  Nemani]{basu2015deepsat}
Saikat Basu, Sangram Ganguly, Supratik Mukhopadhyay, Robert DiBiano, Manohar
  Karki, and Ramakrishna Nemani.
\newblock Deepsat: a learning framework for satellite imagery.
\newblock In \emph{Proceedings of the 23rd SIGSPATIAL international conference
  on advances in geographic information systems}, pages 1--10, 2015.

\bibitem[Helber et~al.(2019)Helber, Bischke, Dengel, and
  Borth]{helber2019eurosat}
Patrick Helber, Benjamin Bischke, Andreas Dengel, and Damian Borth.
\newblock Eurosat: A novel dataset and deep learning benchmark for land use and
  land cover classification.
\newblock \emph{IEEE Journal of Selected Topics in Applied Earth Observations
  and Remote Sensing}, 12\penalty0 (7):\penalty0 2217--2226, 2019.

\bibitem[Demir et~al.(2018)Demir, Koperski, Lindenbaum, Pang, Huang, Basu,
  Hughes, Tuia, and Raska]{demir2018deepglobe}
Ilke Demir, Krzysztof Koperski, David Lindenbaum, Guan Pang, Jing Huang, Saikat
  Basu, Forest Hughes, Devis Tuia, and Ramesh Raska.
\newblock Deepglobe 2018: A challenge to parse the earth through satellite
  images.
\newblock In \emph{2018 IEEE/CVF Conference on Computer Vision and Pattern
  Recognition Workshops (CVPRW)}, pages 172--17209. IEEE, 2018.

\bibitem[Russakovsky et~al.(2015)Russakovsky, Deng, Su, Krause, Satheesh, Ma,
  Huang, Karpathy, Khosla, Bernstein, et~al.]{russakovsky2015imagenet}
Olga Russakovsky, Jia Deng, Hao Su, Jonathan Krause, Sanjeev Satheesh, Sean Ma,
  Zhiheng Huang, Andrej Karpathy, Aditya Khosla, Michael Bernstein, et~al.
\newblock Imagenet large scale visual recognition challenge.
\newblock \emph{International journal of computer vision}, 115\penalty0
  (3):\penalty0 211--252, 2015.

\bibitem[Penatti et~al.(2015)Penatti, Nogueira, and
  Dos~Santos]{penatti2015deep}
Ot{\'a}vio~AB Penatti, Keiller Nogueira, and Jefersson~A Dos~Santos.
\newblock Do deep features generalize from everyday objects to remote sensing
  and aerial scenes domains?
\newblock In \emph{Proceedings of the IEEE conference on computer vision and
  pattern recognition workshops}, pages 44--51, 2015.

\bibitem[Xie et~al.(2016)Xie, Jean, Burke, Lobell, and Ermon]{xie2016transfer}
Michael Xie, Neal Jean, Marshall Burke, David Lobell, and Stefano Ermon.
\newblock Transfer learning from deep features for remote sensing and poverty
  mapping.
\newblock In \emph{Thirtieth AAAI Conference on Artificial Intelligence}, 2016.

\bibitem[Zhou et~al.(2018{\natexlab{a}})Zhou, Zhang, and Wu]{zhou2018d}
Lichen Zhou, Chuang Zhang, and Ming Wu.
\newblock D-linknet: Linknet with pretrained encoder and dilated convolution
  for high resolution satellite imagery road extraction.
\newblock In \emph{CVPR Workshops}, pages 182--186, 2018{\natexlab{a}}.

\bibitem[Zhou et~al.(2018{\natexlab{b}})Zhou, Zheng, Ye, Pu, and
  Sun]{zhou2018satellite}
Zhao Zhou, Yingbin Zheng, Hao Ye, Jian Pu, and Gufei Sun.
\newblock Satellite image scene classification via convnet with context
  aggregation.
\newblock In \emph{Pacific Rim Conference on Multimedia}, pages 329--339.
  Springer, 2018{\natexlab{b}}.

\bibitem[Uzkent et~al.(2019)Uzkent, Sheehan, Meng, Tang, Burke, Lobell, and
  Ermon]{uzkent2019learning}
Burak Uzkent, Evan Sheehan, Chenlin Meng, Zhongyi Tang, Marshall Burke, David
  Lobell, and Stefano Ermon.
\newblock Learning to interpret satellite images in global scale using
  wikipedia.
\newblock \emph{arXiv preprint arXiv:1905.02506}, 2019.

\bibitem[Jean et~al.(2019)Jean, Wang, Samar, Azzari, Lobell, and
  Ermon]{jean2019tile2vec}
Neal Jean, Sherrie Wang, Anshul Samar, George Azzari, David Lobell, and Stefano
  Ermon.
\newblock Tile2vec: Unsupervised representation learning for spatially
  distributed data.
\newblock In \emph{Proceedings of the AAAI Conference on Artificial
  Intelligence}, volume~33, pages 3967--3974, 2019.

\bibitem[Sidek and Quadri(2012)]{sidek2012review}
Othman Sidek and SA~Quadri.
\newblock A review of data fusion models and systems.
\newblock \emph{International Journal of Image and Data Fusion}, 3\penalty0
  (1):\penalty0 3--21, 2012.

\bibitem[Castagno and Atkins(2018)]{castagno2018roof}
Jeremy Castagno and Ella Atkins.
\newblock Roof shape classification from lidar and satellite image data fusion
  using supervised learning.
\newblock \emph{Sensors}, 18\penalty0 (11):\penalty0 3960, 2018.

\bibitem[Correa et~al.(2015)Correa, Bovolo, and Bruzzone]{correa2015vhr}
Yady Tatiana~Solano Correa, Francesca Bovolo, and Lorenzo Bruzzone.
\newblock Vhr time-series generation by prediction and fusion of multi-sensor
  images.
\newblock In \emph{2015 IEEE International Geoscience and Remote Sensing
  Symposium (IGARSS)}, pages 3298--3301. IEEE, 2015.

\bibitem[Schmitt and Zhu(2016)]{schmitt2016data}
Michael Schmitt and Xiao~Xiang Zhu.
\newblock Data fusion and remote sensing: An ever-growing relationship.
\newblock \emph{IEEE Geoscience and Remote Sensing Magazine}, 4\penalty0
  (4):\penalty0 6--23, 2016.

\bibitem[Gao et~al.(2006)Gao, Masek, Schwaller, and Hall]{gao2006blending}
Feng Gao, Jeff Masek, Matt Schwaller, and Forrest Hall.
\newblock On the blending of the landsat and modis surface reflectance:
  Predicting daily landsat surface reflectance.
\newblock \emph{IEEE Transactions on Geoscience and Remote sensing},
  44\penalty0 (8):\penalty0 2207--2218, 2006.

\bibitem[Benedetti et~al.(2018)Benedetti, Raffaele, Kenji, Pensa, Stephane,
  Ienco, et~al.]{benedetti2018m3fusion}
Paola Benedetti, Gaetano Raffaele, Os{\'e} Kenji, Ruggero~Gaetano Pensa, Dupuy
  Stephane, Dino Ienco, et~al.
\newblock M$^3$fusion: Un mod{\`e}le d’apprentissage profond pour la fusion
  de donn{\'e}es satellitaires
  multi-$\{$Echelles/Modalit{\'e}s/Temporelles$\}$.
\newblock In \emph{Conf{\'e}rence Fran{\c{c}}aise de Photogramm{\'e}trie et de
  T{\'e}l{\'e}d{\'e}tection CFPT 2018}, pages 1--8, 2018.

\bibitem[Liggins~II et~al.(2017)Liggins~II, Hall, and
  Llinas]{liggins2017handbook}
Martin Liggins~II, David Hall, and James Llinas.
\newblock \emph{Handbook of multisensor data fusion: theory and practice}.
\newblock CRC press, 2017.

\bibitem[Harrison and Jupp(1989)]{harrison1989introduction}
Barbara~Anne Harrison and David Laurence~Barry Jupp.
\newblock \emph{Introduction to remotely sensed data: Part one of the
  microBrian resource manual}.
\newblock East Melbourne, Vic: CSIRO Publications, 1989.

\bibitem[Gillespie et~al.(1986)Gillespie, Kahle, and
  Walker]{gillespie1986color}
Alan~R Gillespie, Anne~B Kahle, and Richard~E Walker.
\newblock Color enhancement of highly correlated images. i. decorrelation and
  hsi contrast stretches.
\newblock \emph{Remote Sensing of Environment}, 20\penalty0 (3):\penalty0
  209--235, 1986.

\bibitem[CARPER et~al.(1990)CARPER, LILLESAND, and KIEFER]{carper1990use}
WJOSEPH CARPER, THOMASM LILLESAND, and RALPHW KIEFER.
\newblock The use of intensity-hue-saturation transformations for merging spot
  panchromatic and multispectral image data.
\newblock \emph{Photogrammetric Engineering and remote sensing}, 56\penalty0
  (4):\penalty0 459--467, 1990.

\bibitem[Ranchin and Wald(2000)]{ranchin2000fusion}
Thierry Ranchin and Lucien Wald.
\newblock Fusion of high spatial and spectral resolution images: The arsis
  concept and its implementation.
\newblock 2000.

\bibitem[Piao et~al.(2019)Piao, Chen, and Shin]{piao2019new}
Jingchun Piao, Yunfan Chen, and Hyunchul Shin.
\newblock A new deep learning based multi-spectral image fusion method.
\newblock \emph{Entropy}, 21\penalty0 (6):\penalty0 570, 2019.

\bibitem[Li et~al.(2018)Li, Wu, and Kittler]{li2018infrared}
Hui Li, Xiao-Jun Wu, and Josef Kittler.
\newblock Infrared and visible image fusion using a deep learning framework.
\newblock In \emph{2018 24th International Conference on Pattern Recognition
  (ICPR)}, pages 2705--2710. IEEE, 2018.

\bibitem[Daughtry et~al.(2000)Daughtry, Walthall, Kim, De~Colstoun, and
  McMurtrey~Iii]{daughtry2000estimating}
CST Daughtry, CL~Walthall, MS~Kim, E~Brown De~Colstoun, and JE~McMurtrey~Iii.
\newblock Estimating corn leaf chlorophyll concentration from leaf and canopy
  reflectance.
\newblock \emph{Remote sensing of Environment}, 74\penalty0 (2):\penalty0
  229--239, 2000.

\bibitem[Sripada et~al.(2005)Sripada, Heiniger, White, and
  Weisz]{sripada2005aerial}
Ravi~P Sripada, Ronnie~W Heiniger, Jeffrey~G White, and Randy Weisz.
\newblock Aerial color infrared photography for determining late-season
  nitrogen requirements in corn.
\newblock \emph{Agronomy Journal}, 97\penalty0 (5):\penalty0 1443--1451, 2005.

\bibitem[Pearson(1972)]{pearson1972remote}
Robert~Lawrence Pearson.
\newblock Remote mapping of standing crop biomass for estimation of the
  productivity of the shortgrass prairie.
\newblock In \emph{Eighth International Symposium on Remote Sensing of
  Enviroment}, pages 1357--1381. University of Michigan, 1972.

\bibitem[Baret and Guyot(1991)]{baret1991potentials}
Fred Baret and Gerard Guyot.
\newblock Potentials and limits of vegetation indices for lai and apar
  assessment.
\newblock \emph{Remote sensing of environment}, 35\penalty0 (2-3):\penalty0
  161--173, 1991.

\bibitem[Ioffe and Szegedy(2015)]{ioffe2015batch}
Sergey Ioffe and Christian Szegedy.
\newblock Batch normalization: Accelerating deep network training by reducing
  internal covariate shift.
\newblock \emph{arXiv preprint arXiv:1502.03167}, 2015.

\bibitem[Ba et~al.(2016)Ba, Kiros, and Hinton]{ba2016layer}
Jimmy~Lei Ba, Jamie~Ryan Kiros, and Geoffrey~E Hinton.
\newblock Layer normalization.
\newblock \emph{arXiv preprint arXiv:1607.06450}, 2016.

\bibitem[Ulyanov et~al.(2016)Ulyanov, Vedaldi, and
  Lempitsky]{ulyanov2016instance}
Dmitry Ulyanov, Andrea Vedaldi, and Victor Lempitsky.
\newblock Instance normalization: The missing ingredient for fast stylization.
\newblock \emph{arXiv preprint arXiv:1607.08022}, 2016.

\bibitem[Wu and He(2018)]{wu2018group}
Yuxin Wu and Kaiming He.
\newblock Group normalization.
\newblock In \emph{Proceedings of the European Conference on Computer Vision
  (ECCV)}, pages 3--19, 2018.

\bibitem[Nam and Kim(2018)]{nam2018batch}
Hyeonseob Nam and Hyo-Eun Kim.
\newblock Batch-instance normalization for adaptively style-invariant neural
  networks.
\newblock In \emph{Advances in Neural Information Processing Systems}, pages
  2558--2567, 2018.

\bibitem[Tan and Le(2019)]{tan2019efficientnet}
Mingxing Tan and Quoc~V Le.
\newblock Efficientnet: Rethinking model scaling for convolutional neural
  networks.
\newblock \emph{arXiv preprint arXiv:1905.11946}, 2019.

\bibitem[He et~al.(2016)He, Zhang, Ren, and Sun]{he2016deep}
Kaiming He, Xiangyu Zhang, Shaoqing Ren, and Jian Sun.
\newblock Deep residual learning for image recognition.
\newblock In \emph{Proceedings of the IEEE conference on computer vision and
  pattern recognition}, pages 770--778, 2016.

\bibitem[Lin et~al.(2017)Lin, Goyal, Girshick, He, and
  Doll{\'a}r]{lin2017focal}
Tsung-Yi Lin, Priya Goyal, Ross Girshick, Kaiming He, and Piotr Doll{\'a}r.
\newblock Focal loss for dense object detection.
\newblock In \emph{Proceedings of the IEEE international conference on computer
  vision}, pages 2980--2988, 2017.

\bibitem[Sudre et~al.(2017)Sudre, Li, Vercauteren, Ourselin, and
  Cardoso]{sudre2017generalised}
Carole~H Sudre, Wenqi Li, Tom Vercauteren, Sebastien Ourselin, and M~Jorge
  Cardoso.
\newblock Generalised dice overlap as a deep learning loss function for highly
  unbalanced segmentations.
\newblock In \emph{Deep learning in medical image analysis and multimodal
  learning for clinical decision support}, pages 240--248. Springer, 2017.

\bibitem[Kingma and Ba(2014)]{kingma2014adam}
Diederik~P Kingma and Jimmy Ba.
\newblock Adam: A method for stochastic optimization.
\newblock \emph{arXiv preprint arXiv:1412.6980}, 2014.

\bibitem[Loshchilov and Hutter(2016)]{loshchilov2016sgdr}
Ilya Loshchilov and Frank Hutter.
\newblock Sgdr: Stochastic gradient descent with warm restarts.
\newblock \emph{arXiv preprint arXiv:1608.03983}, 2016.

\end{thebibliography}
}

\end{document}